\documentclass[10pt]{article}
\usepackage[a4paper,top=2.54cm,bottom=4.94cm,left=1.93cm,right=1.93cm]{geometry}
\usepackage{booktabs}
\usepackage{longtable}
\usepackage{array}
\usepackage{xurl}
\usepackage{hyperref}
\usepackage{xcolor}
\usepackage{soul}
\usepackage{amsmath}
\usepackage{graphicx}
\usepackage{enumitem}
\usepackage{titlesec}
\usepackage[labelsep=period,font=small]{caption}

\title{CLAP: Closed-Loop Training, Evaluation, and Release Control for Domain Agent Post-training}
\author{{\scriptsize\setlength{\tabcolsep}{2pt}\begin{tabular}{ccc}
Fangfei Li & Chenyang Zhao & Long Wang\\
MatrixOrigin & MatrixOrigin & MatrixOrigin\\
China & China & China\\
lifangfei@matrixorigin.cn & zhaochenyang@matrixorigin.cn & wanglong@matrixorigin.io\\[3pt]
Feng Tian & Zhiyue Zheng & Lv Guo$^{*}$\\
MatrixOrigin & MatrixOrigin & MatrixOrigin\\
USA & China & China\\
tianfeng@matrixorigin.cn & zhengzhiyue@matrixorigin.cn & guolv@matrixorigin.cn$^{*}$
\end{tabular}}}
\date{}

\setlist{nosep,leftmargin=*}
\titlespacing*{\section}{0pt}{3pt}{1pt}
\titlespacing*{\subsection}{0pt}{2pt}{0pt}
\titleformat{\section}{\centering\bfseries\fontsize{11}{13}\selectfont}{\thesection.}{0.5em}{}
\titleformat{\subsection}{\bfseries\fontsize{10}{12}\selectfont}{\thesubsection}{0.5em}{}
\makeatletter
\renewcommand{\maketitle}{%
  \begin{center}
    {\bfseries\fontsize{16}{18}\selectfont \@title\par}
    \vspace{4pt}
    {\fontsize{12}{14}\selectfont \@author\par}
  \end{center}
  \vspace{3pt}
}
\makeatother
\newcommand{\citeup}[1]{\textsuperscript{[#1]}}
\sethlcolor{yellow}
\newcommand{\rev}[1]{#1}
\newcommand{\revblock}[1]{#1}

\begin{document}
\maketitle

\begin{abstract}
Domain agents often face noisy business data, uncertain post-training gains, offline/application mismatch, and adapter-release risk. This paper presents CLAP (Closed-Loop Agent Post-training), a closed-loop method that converts business data into structured SFT samples, decision-preference samples, holdout sets, risk diagnostics, and release-gate records. CLAP combines data validation, target/evidence normalization, reward/KL diagnosis, offline gates, and application-chain replay to decide whether an adapter is suitable for the target application chain. On five anonymized manufacturing-scenario batches, QLoRA-style LoRA-SFT yields modest average gains: overall score increases by 0.0098, pass rate by 0.0240, and evidence accuracy by 0.0280, while hallucination and wrong facts decrease. Yet only 3 of 5 batches improve, some batches regress, and GRPO exposes high KL risks. Application-chain replay further shows that RAG is necessary for factual extraction; under the same 3B backbone and 100 replay cases, an application-RAG-oriented LoRA-SFT adapter improves value, core fields, and answer-evidence doc/page matching over base+RAG, but increases latency. These results support managing domain-agent post-training through an integrated data-training-evaluation-release loop rather than relying on training completion or a single offline score.
\end{abstract}


\section{Introduction}
Domain agents are increasingly used for data questioning, contract lookup, document QA, and process assistance. RAG is a common way to ground LLM applications with external evidence.\citeup{1}\citeup{9} LoRA and QLoRA make parameter-efficient domain adaptation practical.\citeup{2}\citeup{5} Instruction tuning, preference optimization, and RLHF further motivate post-training control for application behavior.\citeup{3}\citeup{4} Reasoning-and-acting agents, benchmarked agent behaviors, tool-use agents, and automated post-training pipelines further show that agent adaptation must be evaluated in the target application workflow.\citeup{15}\citeup{16}\citeup{17}\citeup{18} Such systems accumulate domain data and interaction traces, but raw business signals are not directly reusable post-training assets: they often contain noisy fields, inconsistent formats, incomplete evidence, and unstable labeling conventions. Whether a trained adapter improves business capability also requires reproducible evaluation and release gating.

This paper studies how to transform real business data into post-training assets that are trainable, diagnosable, evaluable, and releasable, and how to determine whether a trained adapter is suitable for the target application chain. Unlike purely algorithmic work, we focus on a system method and experiments on real business data. Using data from an anonymized manufacturing scenario as the case study, we design and validate the CLAP post-training workflow, including data construction, SFT/GRPO training, reward/KL risk checks, holdout evaluation, adapter gating, and application-chain replay.

The main contributions are fourfold: we propose a closed-loop training, evaluation, and release-control workflow for domain agent post-training; design target/evidence normalization for noisy structured extraction targets; conduct base-vs-QLoRA-style LoRA-SFT experiments on five batches of real manufacturing-domain data; and use batch-level regression, GRPO failure-boundary experiments, and application-chain replay to show why domain post-training requires diagnostics, guardrails, application-chain validation, and fallback rather than unconditional promotion.

\section{Related Work and Positioning}
Domain-agent optimization usually combines RAG, SFT, and preference/RL-style training. RAG provides business evidence but does not by itself teach stable output formats or decision preferences.\citeup{1}\citeup{9} SFT and PEFT improve format following, but they can regress when targets and evaluation are misaligned.\citeup{2}\citeup{5} DPO/GRPO-style methods express preferences but are sensitive to reward design, rollout quality, KL drift, and sample distribution.\citeup{4}\citeup{13} \rev{Recent international journal surveys on large language models and retrieval-augmented generation further motivate this governance-oriented positioning.}\citeup{8}\citeup{9} \rev{Surveys on hallucination, parameter-efficient fine-tuning, and human-feedback learning also show that model adaptation must be evaluated together with factuality, risk, and feedback signals.}\citeup{10}\citeup{11}\citeup{12}\citeup{14} CLAP does not propose a new model architecture or claim algorithmic superiority over full-parameter fine-tuning, RAFT, or Self-RAG.\citeup{6}\citeup{7} It addresses a production question: how to convert real business data into training assets, diagnose training risks, gate adapter release, and feed failure cases into the next data and training iteration.

\section{CLAP Method and System Architecture}
\subsection{Task and Asset Definition}
Given a domain business data collection $D$, CLAP converts it into a post-training asset set:
\begin{equation}
A = \{D_{\mathrm{sft}}, D_{\mathrm{grpo}}, D_{\mathrm{eval}}, C, G, R\}.
\end{equation}
where $D_{\mathrm{sft}}$ denotes supervised fine-tuning data, $D_{\mathrm{grpo}}$ denotes preference or decision training data, $D_{\mathrm{eval}}$ denotes holdout evaluation data, $C$ denotes data-construction and training configuration, $G$ denotes gate rules, and $R$ denotes training and evaluation records. The system goal is not to maximize a single training loss, but to make adapter release conditional on gate constraints:
\begin{equation}
\mathrm{promote}(M') \Leftarrow
\mathrm{quality}(M') \geq \mathrm{quality}(M) \land \mathrm{risk}(M') \leq \tau.
\end{equation}
where $M$ is the base model or previous adapter, $M'$ is the candidate adapter, and quality/risk are measured by holdout metrics, factual errors, KL, latency, and regressions. Otherwise, the candidate is held, canaried, or rejected.

\subsection{Training Admission and Release Gating}
CLAP decomposes training into two gates. The pre-training gate checks train/eval overlap, target/evidence verifiability, GRPO reward variance, and chosen/rejected ordering. The post-training gate compares the candidate adapter against the baseline using a configurable policy over quality, risk, latency, and critical regressions.

For SFT, this paper focuses on field, value, and evidence consistency in structured extraction. For GRPO, the method requires pre-training admission and runtime guardrails to be used together, so release readiness is not inferred from static reward distribution alone; the concrete risks are examined in the experiments.

Unlike a single-point gate that checks one final score, CLAP organizes release control across asset validation, training admission, runtime diagnosis, holdout regression checks, application-chain replay, and release/fallback records. Its decision object is a traceable evidence bundle, including dataset version, training configuration, adapter artifact, evaluation record, and regressed samples. Thus adapter release becomes traceable risk decision-making rather than one-time scoring.

\subsection{Method Modules and Decision Logic}
CLAP can be summarized as a control-oriented closed-loop post-training method:
\begin{equation}
\mathrm{assetize} \rightarrow \mathrm{materialize} \rightarrow \mathrm{admit} \rightarrow \mathrm{train} \rightarrow \mathrm{evaluate} \rightarrow \mathrm{gate} \rightarrow \mathrm{app\ validate} \rightarrow \mathrm{release/fallback}.
\end{equation}
Table~\ref{tab:method-modules-en} lists the inputs, outputs, and decision roles of each module.

\begin{table}[htbp]
\footnotesize
\begin{center}
\begin{tabular}{p{0.22\textwidth}p{0.34\textwidth}p{0.34\textwidth}}
\toprule
Module & Input & Output and decision role \\
\midrule
Data assetization & Raw business tables, RAG chunks, feedback signals & Dataset commits, sample indices, and holdout candidates for traceable assets \\
Task materialization & Data assets and task profiles & Structured SFT data, decision GRPO data, and train/eval splits \\
Training admission & Sample quality, overlap, reward distribution & Admit / hold / SFT-only decisions before expensive training \\
Risk diagnosis & Training logs, rollout, KL, reward, length & Early stop events, recipe warnings, and guardrail records \\
Holdout evaluation & Candidate adapter, baseline, holdout set & Structured/decision metrics, regressed samples, and error types \\
Release gate & Evaluation results, latency, error rate, baseline delta & Promote / canary / hold / reject / rollback records \\
Application validation & Target application chain, RAG context, candidate adapter & Validate the complementarity between adapter behavior and retrieved evidence \\
\bottomrule
\end{tabular}
\end{center}
\caption{CLAP method modules, inputs, outputs, and decision roles.}
\label{tab:method-modules-en}
\end{table}

The main decision rules are compact: nonzero train/eval overlap blocks formal training; unverifiable targets or evidence go to data repair; weak GRPO reward gaps or ordering quality lead to SFT-only or hold; runtime KL, reward-drop, or length anomalies stop training; holdout regression or latency/error violations prevent promotion. If adapter-only replay cannot recover factual values or evidence, the adapter is treated as a structured-output enhancement component, not a RAG replacement.

\subsection{System Architecture}
CLAP is implemented as a post-training assetization and release-control system rather than a single training script. It separates data flow from control flow through four layers: asset, materialization, training control, and evaluation/release. Data flow converts raw business records and RAG chunks into dataset commits, sample indices, task-specific train/eval files, and adapter artifacts; control flow records admission, recipes, KL/reward/length risks, baseline comparisons, and promote/canary/hold/reject/rollback decisions. Versioned data management or Git-like database infrastructure can host dataset versions, experiment snapshots, and rollback points, but this paper does not evaluate the database layer itself. Failure cases are fed back to update data-construction rules, reward targets, and gate thresholds in the next iteration.

\begin{figure}[htbp]
\centering
\includegraphics[width=0.82\linewidth]{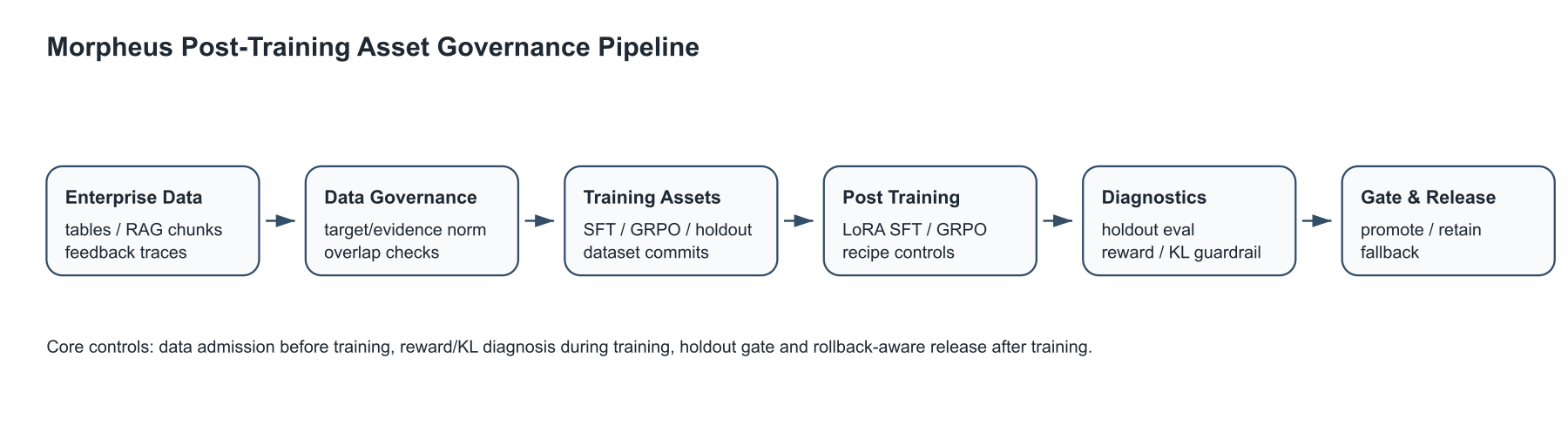}
\caption{CLAP post-training control loop.}
\label{fig:pipeline-en}
\end{figure}

The agents studied here are business-assistant agents for data questioning, contract lookup, and RAG-based QA. We do not claim a full self-evolving loop based on online trajectories; the experiments focus on offline post-training asset construction, evaluation, and pre-release replay.

\section{Data and Experimental Setup}
\subsection{Structured SFT Data}
Structured SFT samples are built from RAG document chunks from the anonymized manufacturer. Each sample contains source material, target constraints, and expected JSON output with fields including entity, metric, period, value, unit, and evidence. In earlier data construction, some metric labels contained numeric values, such as ``6,076.97, proportion of total raw material procurement''. Such value-bearing metric labels make the target unstable and affect holdout evaluation.

Therefore, v3 data construction introduces target/evidence normalization: it strips leading numeric values from metric labels, requires expected values and key metric terms to be verifiable in the input material, and removes ID or input overlap between train and eval splits.

\subsection{Decision GRPO Data}
Decision samples are built from business records such as material backlog and quality-loss reports. GRPO data is organized into groups. Each group contains a grounded chosen candidate, near-miss candidates, and rejected candidates. The reward design focuses on whether the action comes from the original handling suggestion, whether the reason cites the original cause fields, and whether the risk focus follows business rules.

In this paper, GRPO is not used as the main positive result. It is used as evidence for failure boundaries and the necessity of guardrails.

\subsection{Application-chain Evaluation Data}
To examine adapter boundaries, we replay batch\_05 as pre-release application-chain validation. All replay groups use the same 100 cases. The new controlled application-side ABCD experiment uses the same local 3B backbone: A is 3B base+RAG, B is 3B application-RAG-oriented LoRA-SFT+RAG, C is 3B base-only, and D is 3B LoRA-SFT-only. This design separates two questions: whether RAG is necessary for factual extraction, and whether an application-oriented LoRA-SFT adapter can improve synthesis over retrieved evidence under the same 3B backbone. The earlier structured-SFT adapter is retained only as a reference case for training-objective/application-objective mismatch, not as a main table group.

\subsection{Data Version and Configuration}
The main experiment uses \texttt{enterprise\_manufacturing\_v3}. It contains five batches of structured SFT data, each with 100 holdout samples. After overlap cleanup, the training split contains 359--360 samples per batch.

The SFT configuration uses learning rate \texttt{2e-5}, one epoch, per-device train batch size \texttt{2}, gradient accumulation steps \texttt{8}, and evaluation profile \texttt{manufacturing\_structured}. The SFT condition uses Unsloth/PEFT with a 4-bit loaded base model and LoRA adapter training. The LoRA rank is 16 and alpha is 32, so we describe it as QLoRA-style LoRA-SFT. It is not full-parameter fine-tuning and does not constitute an algorithmic comparison against RAFT or Self-RAG.

\subsection{Evaluation Metrics}
Offline structured evaluation metrics include overall score, pass rate, field accuracy, value accuracy, evidence accuracy, hallucination rate, target match, wrong-fact rate, and p95 latency. Hallucination and wrong-fact rates are better when lower; other quality metrics are better when higher. Application-chain metrics include answer JSON parseability, value exact match, period exact match, all-core-fields exact match, evidence chunk/doc/page exact match, retrieved target chunk, and retrieved target doc/page.

\section{Results}
\subsection{Component-level Interpretation}
This section answers three CLAP decisions: whether data is trainable and evaluable, whether gains hold across batches, and whether a candidate adapter suits the target application chain. We therefore report data quality, batch-level regressions, GRPO runtime risk, and ABCD application-chain replay rather than only average scores.

Data quality validation confirms the formal experiment conditions: for all five v3 batches, holdout size is 100, training size is 359--360, and ID overlap, input overlap, noisy metric labels, and missing evidence are all zero after cleanup.

Table~\ref{tab:module-evidence-en} maps CLAP modules to the observed risk types in this paper. We do not interpret these results as strict ablation optimality of every module. Instead, they show that without the corresponding module, a specific class of post-training risk may remain invisible in the data, training, or application chain.

\begin{table}[htbp]
\scriptsize
\centering
\caption{Mapping CLAP modules to observed risks and evidence.}
\label{tab:module-evidence-en}
\setlength{\tabcolsep}{3pt}
\begin{tabular}{@{}p{0.16\textwidth}p{0.29\textwidth}p{0.28\textwidth}p{0.18\textwidth}@{}}
\toprule
Module & Risk intercepted & Evidence in this paper & Decision value \\
\midrule
Data assetization & Noisy labels, unverifiable evidence, train/eval leakage & Zero overlap, noisy metrics, and missing evidence in v3 & Trainable and evaluable data \\
Task materialization & Mixed SFT/GRPO task semantics & Structured SFT is separated from decision GRPO & Avoid objective contamination \\
Training admission & Static samples look trainable but are unsafe for RL & GRPO rank-pass succeeds while KL later becomes abnormal & Allow SFT-only or hold \\
Runtime diagnosis & KL, reward, and length dynamics become abnormal & All five GRPO batches expose high KL risk & Block risky adapters \\
Holdout gate & Average score hides batch-level regression & batch\_01/02 regressions are exposed separately & Canary, hold, or retrain \\
App validation & Offline gain does not imply application-chain gain & Application LoRA+RAG improves core and evidence but increases latency; no-RAG still fails & Promote, canary, or fallback \\
\bottomrule
\end{tabular}
\end{table}

\subsection{Main Base-vs-SFT Results}

\begin{table}[htbp]
\footnotesize
\centering
\caption{Average structured extraction results over five v3 batches. The arrow marks the better direction.}
\label{tab:sft-mean-en}
\begin{tabular}{llrrr}
\toprule
Group & Metric & Base & SFT & Delta \\
\midrule
Overall & Overall ($\uparrow$) & 0.8371 & 0.8469 & +0.0098 \\
Overall & Pass rate ($\uparrow$) & 0.5620 & 0.5860 & +0.0240 \\
Extraction & Field accuracy ($\uparrow$) & 0.8440 & 0.8456 & +0.0016 \\
Extraction & Value accuracy ($\uparrow$) & 0.8980 & 0.9060 & +0.0080 \\
Evidence & Evidence accuracy ($\uparrow$) & 0.6120 & 0.6400 & +0.0280 \\
Evidence & Target match ($\uparrow$) & 0.8933 & 0.8960 & +0.0027 \\
Risk & Hallucination ($\downarrow$) & 0.0260 & 0.0180 & -0.0080 \\
Risk & Wrong fact ($\downarrow$) & 0.0800 & 0.0740 & -0.0060 \\
Latency & P95 latency ($\downarrow$, ms) & 3082 & 2785 & -297 \\
\bottomrule
\end{tabular}
\end{table}

As shown in Table~\ref{tab:sft-mean-en}, SFT improves overall score on 3 of 5 batches and yields small mean gains in overall, pass rate, value, evidence, and target match while reducing hallucination and wrong facts. The clearest gain is evidence-related: evidence accuracy improves by 0.0280 and hallucination decreases from 0.0260 to 0.0180. P95 latency decreases by about 297ms, but serving-path variance prevents treating it as a standalone stable gain. Per-batch overall deltas are -0.0101, -0.0006, +0.0076, +0.0183, and +0.0336 for batch\_01 to batch\_05. The first two batches show that average improvement alone cannot imply every business subdomain is releasable; CLAP turns such regressions into hold, repair, rematerialization, or canary decisions.

\subsection{Risk Boundary and Application-chain Ablation Replay}
Static group-aware admission passes all five GRPO batches, with rank-pass rate equal to 1.0, reward margin mean around 0.224--0.226, and reward std around 0.332--0.334. However, actual GRPO training remains highly sensitive to recipe choices. Table~\ref{tab:risk-app-en} summarizes KL risk and application-chain replay. KL here is a training-log risk signal for relative diagnosis and gating, not a universal cross-model threshold.

\begin{table}[htbp]
\footnotesize
\centering
\caption{GRPO risk diagnosis and application-chain replay.}
\label{tab:risk-app-en}
\begin{tabular}{lrrrrll}
\toprule
\multicolumn{7}{l}{Panel A. GRPO runtime risk. Lower KL is safer.} \\
\midrule
Batch & KL peak & Rolling KL & Reward$_0$ & Reward$_T$ & Diagnosis & Gate \\
\midrule
01 & 2127.8 & 1064.3 & 0.62 & 0.43 & unstable & hold \\
02 & 168.2 & 87.2 & 0.14 & 0.42 & unstable & hold \\
03 & 6427.9 & 3245.4 & 0.52 & 0.48 & unstable & hold \\
04 & 646.8 & 323.8 & 0.58 & 0.40 & unstable & hold \\
05 & 6712.0 & 3816.6 & 0.49 & 0.65 & unstable & hold \\
\bottomrule
\end{tabular}

\vspace{2pt}
\resizebox{\textwidth}{!}{%
\begin{tabular}{llllrrrrrrrrrr}
\toprule
\multicolumn{14}{l}{Panel B. Application-chain replay. Higher quality is better; lower P95 is better.} \\
\midrule
ID & Backbone & RAG & Adapter & JSON & Value & Entity & Metric & Period & Unit & Core & AnsEv & RetDP & P95 \\
A & 3B & yes & no & 0.98 & 0.83 & 0.98 & 0.98 & 0.97 & 0.86 & 0.71 & 0.25 & 0.98 & 3799 \\
B & 3B & yes & App-LoRA & 1.00 & 0.85 & 1.00 & 1.00 & 0.99 & 0.93 & 0.78 & 0.45 & 0.98 & 4582 \\
C & 3B & no & no & 1.00 & 0.00 & 1.00 & 1.00 & 1.00 & 0.92 & 0.00 & 0.02 & 0.00 & 1380 \\
D & 3B & no & App-LoRA & 1.00 & 0.00 & 1.00 & 1.00 & 1.00 & 0.98 & 0.00 & 0.03 & 0.00 & 1599 \\
\bottomrule
\end{tabular}
}
\end{table}

Panel A shows that static reward distribution is necessary but insufficient: all five batches pass static admission, yet runtime training produces extreme KL peaks. Even batch\_05, whose Reward$_T$ exceeds Reward$_0$, is held for KL risk, so reward or rank-pass alone overestimates GRPO release readiness. Panel B uses the same 3B backbone; AnsEv is answer-evidence exactness and RetDP is retrieval hit rate. C/D obtain zero value and core accuracy, confirming that RAG is necessary and the adapter does not replace retrieval. With RAG, the application-oriented LoRA-SFT adapter improves value (0.85 vs 0.83), core fields (0.78 vs 0.71), and answer-evidence doc/page matching (0.45 vs 0.25), but increases P95 latency (4582ms vs 3799ms). Thus application replay turns adapter release into a quality-latency-risk decision rather than unconditional promotion.

\section{Discussion, Limitations, and Conclusion}
CLAP is positioned as a governance-oriented closed-loop method for domain-agent post-training, not as a new training algorithm. The current experiments use QLoRA-style LoRA-SFT and a guarded GRPO boundary study on one anonymized manufacturing scenario, so the conclusions should be interpreted at the method and release-control level rather than as universal algorithmic ranking.

Several limitations remain. First, the data volume is modest and scenario-specific. Second, the application replay still evaluates a bounded pre-release case set rather than full online production traffic. Third, full-parameter fine-tuning, RAFT, Self-RAG, and broader adaptation baselines are outside the present scope. Fourth, GRPO here mainly serves as evidence for failure boundaries and the need for guardrails, not as a stable positive recipe.

Even with these limits, the main result is clear: real business-agent post-training cannot be managed by training completion or a single offline score alone. Small average SFT gains can coexist with per-batch regressions; static GRPO admission can coexist with runtime instability; and offline structured gains do not guarantee application-chain readiness without replay. CLAP therefore provides a practical governance loop for turning business data into post-training assets and for deciding whether a candidate adapter should be promoted, canaried, held, or rolled back.

\end{document}